\documentclass[final]{cvpr}

\usepackage{times}
\usepackage{epsfig}
\usepackage{graphicx}
\usepackage{amsmath}
\usepackage{amssymb}
\usepackage{booktabs}
\usepackage{soul}
\usepackage{siunitx}
\usepackage{subcaption}
\robustify\bfseries

\usepackage[pagebackref=true,breaklinks=true,colorlinks,bookmarks=false]{hyperref}
\usepackage{xcolor}
\usepackage{enumitem}

\newcommand{\KGnote}[1]{{\color{magenta}{\bf KG: }#1}}
\newcommand{\KG}[1]{{\color{black}#1}} %
\newcommand{\KGmon}[1]{{\color{black}#1}} %
\newcommand{\KGcr}[1]{{\color{blue}#1}} %

\newcommand{\cc}[1]{{\color{magenta}#1}}

\newcommand{\CAnote}[1]{{\color{red}{\bf CA: }#1}}
\newcommand{\CA}[1]{{\color{black}#1}} %
\newcommand{\CCA}[1]{{\color{black}#1}} %
\newcommand{\C}[1]{{\color{brown}#1}} %
\newcommand{\CAcr}[1]{{\color{brown}#1}} %

\newcommand{\ZA}[1]{{\color{black}#1}} %
\newcommand{\Z}[1]{{\color{black}#1}} %
\newcommand{\ZZ}[1]{{\color{cyan}#1}} %
\newcommand{\ZAcr}[1]{{\color{cyan}#1}} %

\renewcommand{\KGnote}[1]{{\color{black}{}}}
\renewcommand{\KG}[1]{{\color{black}#1}}
\renewcommand{\KGmon}[1]{{\color{black}#1}}

\renewcommand{\CAnote}[1]{{\color{black}{}}}
\renewcommand{\CA}[1]{{\color{black}#1}}
\renewcommand{\CCA}[1]{{\color{black}#1}}
\renewcommand{\C}[1]{{\color{black}#1}}

\renewcommand{\ZA}[1]{{\color{black}#1}}
\renewcommand{\Z}[1]{{\color{black}#1}}
\renewcommand{\ZZ}[1]{{\color{black}#1}}

\renewcommand{\CAcr}[1]{{\color{black}#1}}
\renewcommand{\ZAcr}[1]{{\color{black}#1}}
\renewcommand{\KGcr}[1]{{\color{black}#1}}
\renewcommand{\cc}[1]{{\color{black}#1}}

\def\cvprPaperID{9940} %
\def\confYear{CVPR 2021}

\begin{document}

\title{Semantic \KG{Audio-Visual} Navigation}%

\author{Changan Chen$^{1,2}$ \hspace{3mm}  Ziad Al-Halah$^{1}$ \hspace{3mm} Kristen Grauman$^{1,2}$\\
$^1$UT Austin \hspace{3mm} $^2$Facebook AI Research
}

\maketitle

\begin{abstract}
\KG{Recent work on audio-visual navigation assumes a constantly-sounding target and restricts the role of audio to signaling the target's position. We introduce \emph{semantic audio-visual navigation}, where objects in the environment make sounds consistent with their semantic \KGcr{meaning} (e.g., toilet flushing, door creaking) and acoustic events are sporadic or short in duration. We propose a transformer-based model to tackle this new semantic AudioGoal task, incorporating an inferred goal descriptor that captures both spatial and semantic properties of the target. Our model’s persistent multimodal memory enables it to reach the goal even long after the acoustic event stops. In support of the new task, we  also expand the SoundSpaces audio simulations to provide semantically grounded sounds for an array of objects in Matterport3D. Our method strongly outperforms existing audio-visual navigation methods by learning to associate semantic, acoustic, and visual cues.\CAcr{\footnote{Project page: \url{http://vision.cs.utexas.edu/projects/semantic-audio-visual-navigation}}}}
\end{abstract}

\vspace*{-0.15in}
\section{Introduction}

An autonomous agent interacts with its environment in a continuous loop of action and perception. 
The agent needs to reason intelligently about all the senses available to it (sight, hearing, proprioception, touch) to select the proper sequence of actions in order to achieve its task.  
\KG{For example, a service robot of the future may} need to locate and fetch an object for a user, \KG{go empty the dishwasher when it stops running, or travel to the front hall upon hearing a guest begin speaking there.}

\KG{Towards such applications, recent progress in visual navigation builds agents that use egocentric vision to travel to a designated point in an unfamiliar environment~\cite{gupta2017cognitive,habitat19iccv,wijmans_dd-ppo_2020,Chaplot2020Learning}, search for a specified object~\cite{zhu-iccv2017,chaplot_object_2020,savinov2018semiparametric,neural-topo}, or explore and map a new space~\cite{ramakrishnan-eccv2018,ramakrishnan_occupancy_2020,chen_learning_2019,Chaplot2020Learning,silvio-scene-memory,Chaplot2020Learning,exploration-ijcv}.}
Limited new work further explores expanding the sensory suite of the \KG{navigating} agent to include hearing as well.  In particular, %
the AudioGoal challenge~\CAcr{\cite{chen_audio-visual_2019}} \KGcr{requires an agent to navigate to a sounding target (e.g., a ringing phone) using audio for key directional and distance cues~\cite{chen_audio-visual_2019,gan2019look,chen_waypoints_2020}.}

\begin{figure}[t]
    \centering
    \includegraphics[width=1.\linewidth]{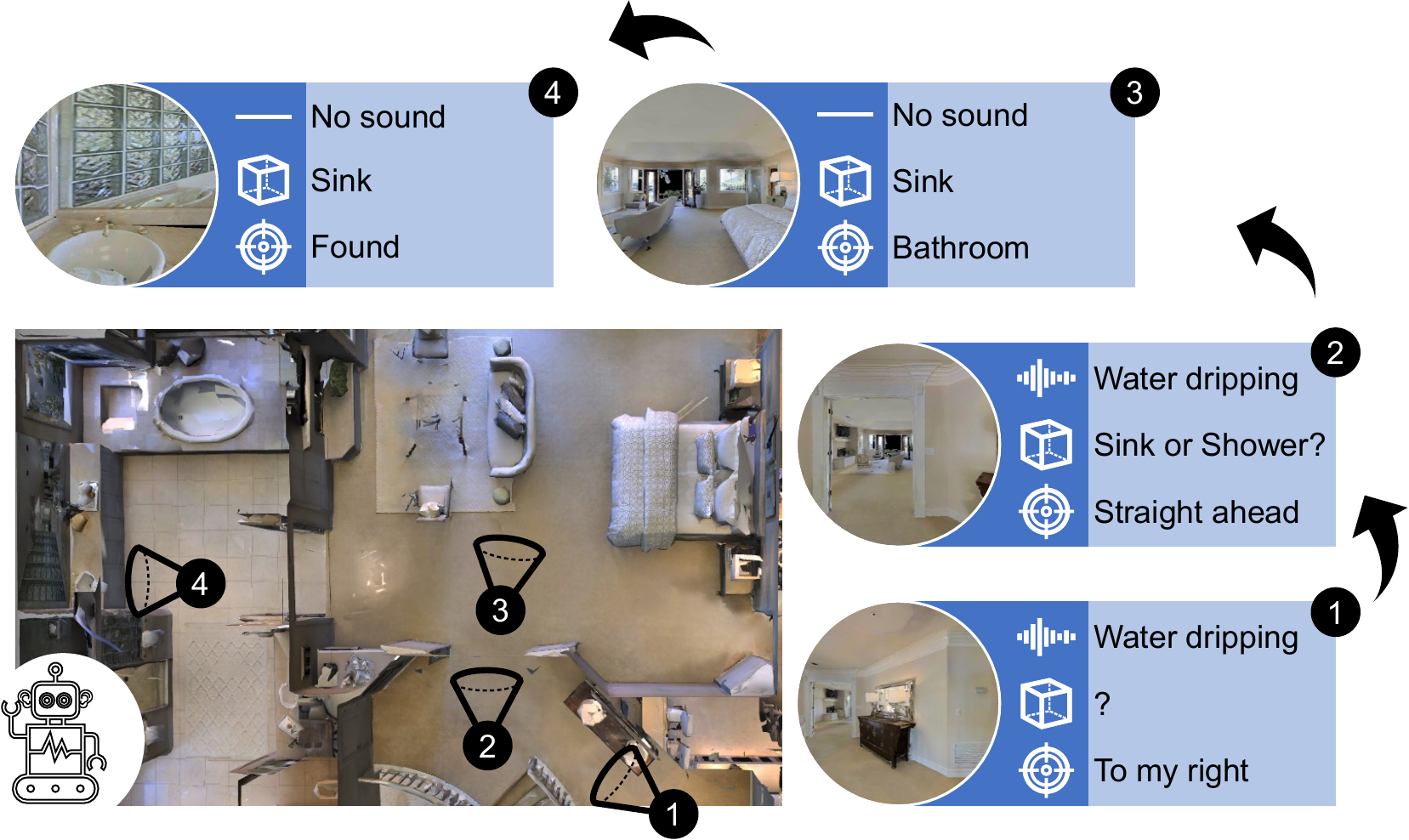}
    \vspace*{-0.3in}
    \caption{Semantic \KG{audio-visual} navigation in 3D environments: an agent must navigate to a sounding object. %
    \KG{Since the sound may stop while the agent searches for the object, the agent is incentivized to learn the association between how objects look and sound, and to build contextual models for where different semantic sounds are more likely to occur (e.g., water \ZAcr{dripping} in the bathroom). %
    }
    }
    \label{fig:intro}
\end{figure}

\KG{While exciting first steps, existing audio-visual navigation work has two key limitations.  First, prior work}
assumes the target object constantly makes a steady repeating sound (e.g., alarm chirping, phone ringing). \KG{While important, this corresponds to a narrow set of targets; in real-world scenarios, an object may emit a sound only briefly or start and stop dynamically.}
\KG{Second, in current models explored in realistic 3D environment simulators, the sound emitting target has neither a visual embodiment nor any semantic context.  Rather, target sound sources are placed arbitrarily in the environment and without relation to the semantics of the scene and objects.  As a result, the role of audio is limited to providing a beacon of sound announcing where the object is.}

\KG{In light of these limitations, we introduce a novel task: \emph{semantic audio-visual navigation}.}
In this task, the agent must navigate to an object \KG{situated contextually in an environment} %
that only makes sound for a certain period of time.  \KG{Semantic audio-visual navigation widens the set of real-world scenarios to include acoustic events of short temporal duration \KGmon{that are semantically grounded in the environment}. 
It offers new learning challenges.}
The agent must learn not only how to associate sounds with visual objects, but also how to leverage the semantic priors of objects (along with any acoustic cues) to reason about where the object is likely located in the scene.
For example, hearing the dishwasher stop running and issue its end of cycle chime should suggest both what visual object to search for as well as the likely paths for finding it, i.e., towards the kitchen rather than the bedroom.   
\KG{Notably, in the proposed task,} the agent is not given any external information about the goal (such as a displacement vector or \KG{name of the object to search for}).
Hence the agent must learn to leverage %
\KG{sporadic}
acoustic cues that may stop at any time as it searches for the source, \KG{inferring what visual object likely emitted the sound even after it is silent.}  See Figure~\ref{fig:intro}. %

To tackle semantic AudioGoal, we introduce a \KG{deep reinforcement learning} model that learns %
the association between how objects look and how they sound. 
\ZAcr{We develop a \emph{goal descriptor} module that allows the agent} %
to hypothesize the goal properties (i.e., location and \KG{object} category) from the received acoustic cues %
\KG{before seeing the target object}.  
Coupled with a transformer, it learns to attend to the previous visual and acoustic observations in its memory---conditioned on the predicted goal descriptor---to navigate to the audio source. 
Furthermore, to support this line of research, we instrument audio-visual simulations for real \KG{scanned} environments such that semantically relevant sounds are attached to semantically relevant objects.

We evaluate our model on 85 large-scale real-world environments with a variety of semantic objects and their sounds.
Our approach outperforms state-of-the-art models in audio-visual navigation with up to \CAcr{an absolute} 8.9\% improvement in SPL. 
Furthermore, our model is robust in handling short acoustic signals \KGmon{emitted by the goal} with varying temporal duration, and compared to the competitors, it %
more often reaches the goal after the acoustic observations end.
\C{In addition, our model maintains good performance in the presence of environment noise (distractor sounds) compared to baseline models.}
\KGcr{Overall, this work shows} the  \KGmon{potential for embodied agents to learn about how objects look and sound through interactions with a 3D environment.}

\vspace*{-0.05in}
\section{Related work}
\vspace*{-0.05in}

\paragraph{Visual navigation.} 
To navigate autonomously, traditionally a robot builds a map via 3D reconstruction  (i.e., SLAM) and then plans a path using the map~\cite{FuentesPacheco2012VisualSL}. 
Recent work instead learns navigation policies directly from egocentric observations~\cite{gupta2017cognitive,savinov2018semiparametric,mishkin2019benchmarking}. 
A popular task is PointGoal navigation, where the goal position is given to the \KGcr{agent}~\cite{gupta2017cognitive,mishkin2019benchmarking,habitat19iccv,wijmans_dd-ppo_2020}.
Alternatively, %
in the ObjectGoal setting, the agent is given an object label rather than the goal location, and 
\KGcr{must navigate}
to the nearest instance of that category (e.g., go to a table)
~\cite{zhu-iccv2017,batra_objectnav_2020,chaplot_object_2020}. 
In contrast to both PointGoal and ObjectGoal, in the proposed setting the agent is not given specific goal information.  Instead, it needs to react to an acoustic event 
to determine what kind of object is sounding and navigate to it. 
Furthermore, unlike ObjectGoal, the agent needs to navigate to the specific object instance that emitted the sound rather than any instance of that category.
Our task represents real-world scenarios where dynamic objects draw the attention of an agent and call it to action (e.g., the sound of a heavy object falling upstairs).

\vspace*{-0.15in}
\paragraph{Audio-visual navigation.}
Recent work leverages audio %
\KGcr{for} the AudioGoal navigation task~\cite{chen_audio-visual_2019,gan2019look}.
In that setup, the agent navigates to a sound-emitting goal using both visual and acoustic observations~\cite{chen_audio-visual_2019,gan2019look,chen_waypoints_2020}. 
As discussed above, prior methods assume the goal is sounding continuously through the episode and that it does not have a visual \KG{embodiment}. 
While suitable for certain events like \emph{fire alarm}, many acoustic events are short and infrequent (e.g., glass breaking, \KG{door slamming}, a person calling for help). 
We consider a generalized setting where the audio signal is only available for a 
\CA{limited period of}
time and the agent must find \ZAcr{the sounding object} using both initial acoustic cues and the goal semantics.
\KG{In addition, we augment the SoundSpaces audio simulations~\cite{chen_audio-visual_2019} for Matterport3D~\cite{Matterport3D} to portray semantically relevant object-level sounds, an advance over the simulations used in prior work~\cite{chen_audio-visual_2019,gan2019look,chen_waypoints_2020}, \KGcr{which} inserted a small set of sounds randomly in the environments without any visual embodiment.}

\vspace{-0.15in}
\paragraph{Memory models for 3D environments.} 
While it is common to use an implicit memory representation in navigation to aggregate observations, e.g., a recurrent network~\cite{mirowski2016learning,habitat19iccv,chen_audio-visual_2019,krantz_beyond_2020,anderson_vision-and-language_2017,mousavian_visual_2018}, other methods leverage explicit map-based memories to record occupancy~\cite{gupta2017cognitive,chen_learning_2019,ramakrishnan_occupancy_2020,Chaplot2020Learning,chen_waypoints_2020,av-map} or object locations~\cite{chaplot_object_2020,cartillier2020semantic}.
To capture long-term \KGcr{dependencies} %
another promising direction is to use a transformer architecture~\cite{vaswani2017attention} to record observations and poses~\cite{silvio-scene-memory}.  We build in this direction and introduce a scene memory transformer that, unlike prior work, \KG{1) is multimodal and 2) leverages} an explicit
 learned \emph{goal descriptor} to attend to the memory.  Our memory model learns audio-visual \KG{associations} between the goal and the observations from the scene, a crucial functionality as we demonstrate in experiments.

\vspace*{-0.15in}
\paragraph{Audio-visual learning in video.}

Work in passive (non-embodied) video analysis also explores 
the link between object appearance and sound.  
This includes audio source separation methods that disentangle sounds based on object appearance~\cite{gao2018eccv,ephrat2018looking,owens2018audio,afouras2018conversation,zhao2018sound,gao20192,gao2019co} \KG{as well as self-supervised video representation learning methods~\cite{korbar2018cooperative,owens2018audiovisual,fernando2017selfsupervised,Gan_2018_CVPR}.}
Our approach also learns how to associate objects with their sounds. 
However, in contrast to previous video approaches, \KG{our approach learns in the context of an agent's interaction with a 3D environment. Namely, our agent learns to associate an object category inferred from audio with its visual representation %
\KG{and contextual scene cues}
at the same time it learns to navigate efficiently.}

\section{Semantic \KG{Audio-Visual} Navigation}\label{sec:task}
We introduce the novel task of semantic audio-visual navigation. 
In this task, the agent is required to navigate in a complex, unmapped environment to find a semantic sounding object---semantic AudioGoal for short.  %
Different from AudioGoal~\cite{chen_audio-visual_2019,gan2019look}, the goal sound \emph{need not be periodic}, has \emph{variable duration}, and is associated with a meaningful \emph{semantic object} (e.g., \ZA{the door creaking is associated with} the apartment's door). %
This setting represents common real world events, %
and as discussed above, poses new challenges for embodied learning. 
Relying on audio perception solely to produce step-by-step actions is not sufficient, since the audio event is relatively short. %
Instead, the agent needs to reason about the category of the sounding object and use both visual and audio perception to predict its location. 

\vspace{-0.15in}
\paragraph{3D environments and simulator.} 
Consistent with the active body of computer vision work on embodied AI done in simulation, and to facilitate reproducibility of our work, we rely on a visually and acoustically realistic simulation platform to model an agent moving in complex 3D environments.  We use SoundSpaces~\cite{chen_audio-visual_2019}, which enables realistic audio rendering of arbitrary sounds for the real-world environment scans in Replica~\cite{straub2019replica} and
 Matterport3D~\cite{Matterport3D}. We use the Matterport environments
 \KG{due} to their greater scale and complexity.  SoundSpaces is Habitat-compatible~\cite{habitat19iccv} and allows rendering arbitrary sounds at any pair of source and receiver (agent) locations on a uniform grid of nodes spaced by 1 m.  Next we explain how we extend this audio data to provide semantically meaningful sounds.

\vspace{-0.15in}
\paragraph{Semantic sounds data collection.}
We use the 21 object categories defined in the ObjectGoal navigation challenge~\cite{batra_objectnav_2020} for Matterport3D environments: chair, table, picture, cabinet, cushion, sofa, bed, chest of drawers, plant, sink, toilet, stool, towel, tv monitor, shower, bathtub, counter, fireplace, gym equipment, seating, and clothes. All of these categories have objects that are visually present in Matterport environments.  
By rendering object-specific sounds at the locations of the Matterport objects, we obtain semantically meaningful and contextual sounds.  
For example, the \ZA{water flush sound will be associated with the toilet in the bathroom}, and the \ZA{crackling fire sound with the fireplace in the living room or the bedroom}.
\cc{We filter out object instances that are not reachable by the navigability graph.}
\CA{The number of object instances for train/val/test is 303/46/80 on average for each object category.}

We consider two types of sound events: object-emitted and object-related.  Object-emitted sounds are generated by the object, e.g., a toilet flushing, whereas object-related sounds are caused by \KGcr{people's} interactions with the object, e.g., food being chopped on the counter.
To provide a variety of sounds, we \KGcr{search a public database \url{freesound.org} by the 21 object names} to
get long copyright-free audio clips \KGcr{per object}.  %
\KG{We split the original clips (average length 81s) evenly into train/val/test clips.}
These splits allow the characteristics of the unheard sounds \ZZ{(i.e., waveforms not heard during training)} to be related to those in the training set, while still preserving natural variations.\footnote{Note that even the same waveform will sound different when rendered in a new environment; the sound received by the agent is a function of not only that waveform but also the environment geometry and the agent's position relative to the source.}
The duration of the acoustic phase in each episode is randomly sampled from a Gaussian of mean 15s and deviation 9s, clipped for a minimum 5s and maximum 500s. If the sampled duration is longer than the length of the audio clip, we replay the clip. \KG{See the Supp.~video for examples.}  

\vspace{-0.15in}
\paragraph{Action space and sensors.} %
The agent's action space is \textit{MoveForward}, \textit{TurnLeft}, \textit{TurnRight}, and \textit{Stop}. The last three actions are always valid, while \textit{MoveForward} only takes effect when the node in front of the agent is reachable from that position \KG{(no collision)}. 
The sensory inputs are egocentric binaural sound \CCA{(two-channel audio waveforms)}, RGB, depth, and the agent's current pose. 

\vspace{-0.16in}
\paragraph{Episode specification and success criterion.} 
An episode of semantic AudioGoal is defined by 1) the scene, 2) the agent start location and rotation, 3) the goal location, 4) \CA{the goal (\KG{object}) category} and 5) the duration of the audio event. 
\CA{In each episode in a given scene, we choose a random object category and a random instance of that category as the goal. 
The agent's \ZA{start} pose is also randomly positioned in the scene.  %
In semantic AudioGoal, the agent has to stop near the particular sounding object instance, not simply any instance of the class. 
This is a stricter success criterion than ObjectGoal~\cite{batra_objectnav_2020}, which judges an episode as successful if the agent stops near any instance of that category.  
We define a set of viewpoints around each object within 1 m of the object's boundary; issuing the \emph{Stop} action at any of these viewpoints is considered a successful \ZA{termination of the episode}.

\begin{figure*}[ht!]
    \centering
    \includegraphics[width=0.80\textwidth]{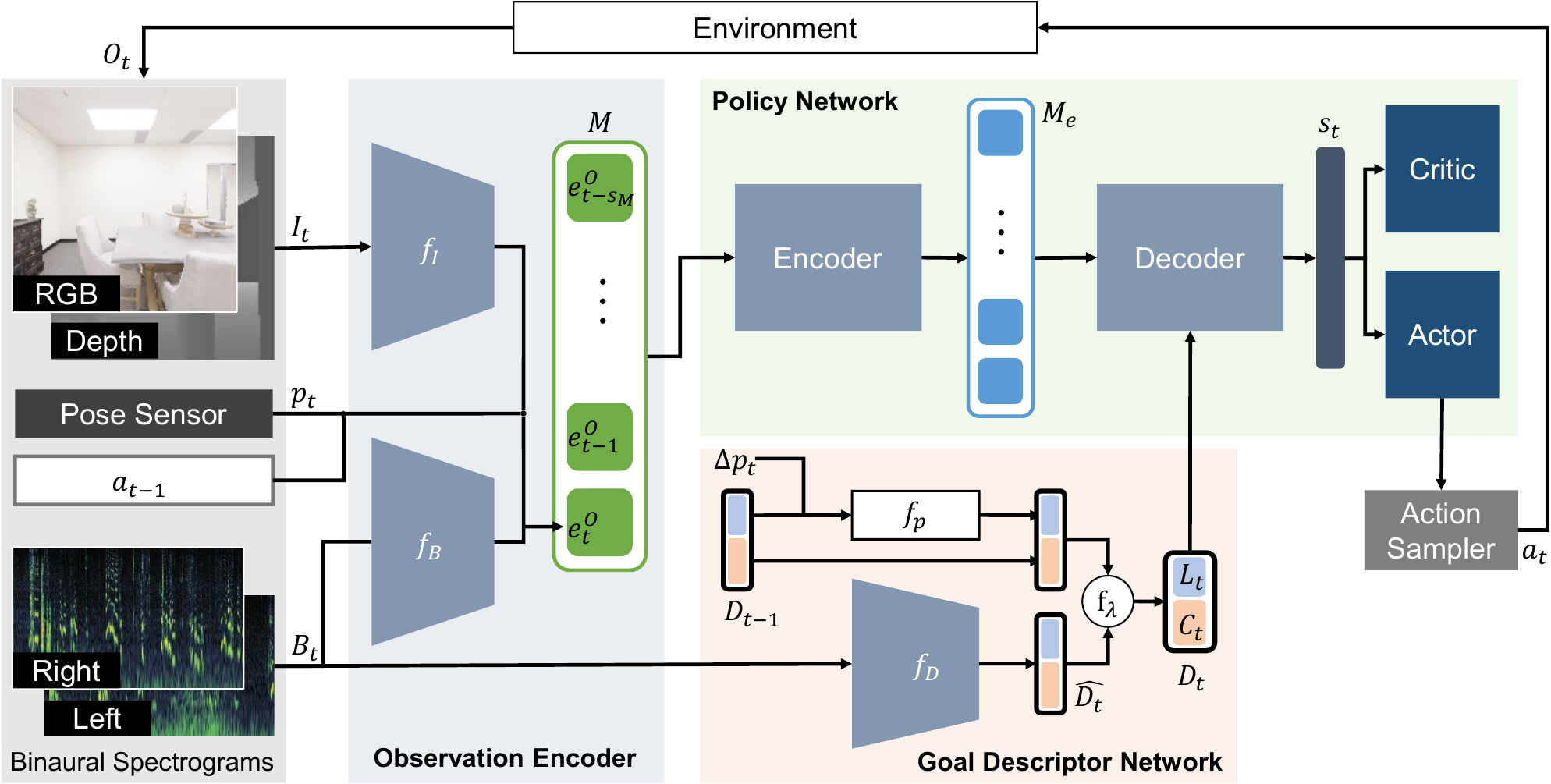}
    \caption{%
    In our model, the agent \CCA{first encodes input observations and stores their features in memory $M$. Then }\ZAcr{our goal descriptor network leverages} the acoustic cues to dynamically infer and update a \emph{goal descriptor} \ZA{$D_t$} of the target object, which contains both location \ZA{$L_t$} and object category \ZA{$C_t$} information about the goal.
    By conditioning the agent's scene memory on the goal descriptor, the learned state representation \ZA{$s_t$} preserves information most relevant to the goal. %
    \ZA{Our transformer-based policy network attends} 
    to the encoded observations %
    \ZA{in $M$ with self-attention to reason about the 3D environment seen so far, and it attends to $M_e$ with $D_t$} 
    to capture possible associations between the hypothesized goal and the visual and acoustic observations \ZA{to predict the state $s_t$.}
    Then, \CA{$s_t$ is fed to \ZA{an actor-}critic network, which predicts the \ZA{next} action \ZA{$a_t$}.}
    \ZA{The agent receives its reward from the environment based on how close to the goal it moves and whether it succeeds in reaching it.} 
    }
    \label{fig:model}
\end{figure*}

\section{Approach}\label{sec:approach} %
\vspace{-0.05in}

We propose \KGcr{SAVi}, a novel model for the \textbf{s}emantic \textbf{a}udio-\textbf{v}isual \textbf{n}av\textbf{i}gation task.
\KGcr{SAVi} uses a persistent multimodal memory along with a transformer model, which,
\ZA{unlike RNN-based architectures (e.g., \cite{chen_audio-visual_2019}) or reactive ones (e.g.,~\cite{gan2019look}), 
\KGcr{can} directly attend to observations with various temporal distances from the current step to locate the goal efficiently.
Furthermore, our model learns to capture goal information from acoustic \CA{events} in an explicit descriptor and uses it to attend to its memory, thus enabling the agent to discover any \CA{spatial and} semantic cues that may help it reach the target faster.}

Our approach has three main components \ZA{(Figure~\ref{fig:model})}:
1) an Observation \ZA{Encoder} that maps the egocentric visual and acoustic observations received by the agent at each step to an embedding space;
2) a Goal \ZA{Descriptor Network} that produces a goal descriptor based on the encoded observations; and
3) a Policy \ZA{Network} that given the encoded observations and the predicted goal descriptor, \CA{extracts a descriptor-conditioned state representation and outputs the action distribution.} 
Next, we describe each module.  \KG{We defer CNN architecture details to Sec.~\ref{sec:training}.}

\subsection{Observation Encoder}\label{sec:obs_encoder} %
At each time step $t$, the agent receives an observation $O_t = (I_t, B_t, p_t, a_{t-1})$, where $I$ is the egocentric visual observation consisting of an RGB \KG{and depth image}; %
$B$ is the received binaural audio waveform represented as a two-channel spectrogram; 
$p$ is the agent pose defined by its location and orientation $(x, y, \theta)$ with respect to its starting pose $p_0$ in the current episode; 
and $a_{t-1}$ is the action taken at the previous time step.

\ZA{Our model} encodes each visual and audio observation with a CNN, $e^I_t=f_I(I_t)$ and $e^B_t=f_B(B_t)$.
Then, the observation $O_t$ encoding is $e^O_t = [e^I_t, e^B_t, p_t, a_{t-1}]$.
\ZA{The model} \CA{stores} the encoding of the observations up to time $t$ in memory $M=\{{e^O_i}: {i=\max\{0, t - s_M\},\ldots,t}\}$ (see Figure~\ref{fig:model} \KG{second column}), \CA{where $s_M$ is the memory size.}

\subsection{Goal Descriptor Network}\label{sec:goal_desc}
As described in Sec.~\ref{sec:task}, the agent does not receive direct information about the goal; rather, it needs to rely solely on its observations to set its own target.
Audio carries rich cues about the target---not only its relative direction and distance from the agent, but also the type of object that may have produced the acoustic event.  
Hence, we leverage the acoustic signal to predict the goal properties, namely its location (spatial) and object category (semantics).
Both properties are crucial for successful navigation.
The estimated goal location gives the agent an idea of %
where to find the goal.
However, since the acoustic event may be short-lived, \KG{and the estimate may be inaccurate}, the agent cannot solely rely on this initial estimate.  Our model thus aims to also leverage the goal semantics in \KGmon{terms of both the object's likely appearance and the scene's visual context.} %

The goal descriptor network is a CNN $f_D$ such that 
$\hat{D}_t=f_D(B_t)$, where $\hat{D}_t$ \ZA{is the step-wise estimate of the descriptor and it} consists of two parts: the current estimate of the goal location $\hat{L}_t = (\Delta x, \Delta y)$ relative to the agent's current pose $p_t$, and its predicted object label $\hat{C}_t$.  
To reduce the impact of noise from a single prediction, the agent \CA{aggregates the current estimate with the previous goal descriptor}
\ZA{$D_t = f_\lambda(\hat{D}_t, D_{t-1}, \Delta p_t) = (1 - \lambda) \hat{D}_t + \lambda f_p(D_{t-1}, \Delta p_t)$}, where $f_p(\cdot)$ \CA{transforms} the previous goal location $\hat{L}_{t-1}$ \CA{based on the last pose change $\Delta p_t$ (the goal label is unaffected by this transformation),}
and $\lambda$ is the weighting factor, which is set to $0.5$ based on validation. 
\CA{When sound stops (\ZA{i.e., the sound intensity} becomes zero), \ZA{the agent maintains its latest estimate $D_t$ by} %
simply transforming the previous descriptor \ZA{based on the pose change $\Delta p_t$} to obtain the current descriptor $D_t = f_p(D_{t-1}, \Delta p_t)$.}

\subsection{Policy Network}\label{sec:policy} %
Our reinforcement learning policy network is based on a transformer architecture.
Using the memory $M$ collected so far in the episode, the transformer proceeds by encoding these observation embeddings with a self-attention mechanism to capture any possible relations among the inputs, \KG{yielding the encoded memory} $M_e=\textrm{Encoder}(M)$.   
Then, using the predicted goal descriptor $D_t$, a decoder network attends to all cells in the encoded memory $M_e$ to calculate the state representation $s_t = \textrm{Decoder}(M_e, D_t)$.
\CCA{
\ZZ{An actor-critic network uses $s_t$ to predict the action distribution and value of the state. The actor and the critic are each}
modelled by single linear layer neural network.} 
Finally, an action sampler samples the next action $a_t$ from this action distribution, determining the agent's next motion in the 3D scene.

\subsection{Training}\label{sec:training}
\CA{To train the goal descriptor network, we 
\KG{generate pairs of ground truth locations and categories from the simulator for the array of training sounds}, and train the prediction network in a supervised fashion. 
For the category prediction \KG{portion}, we find off-policy training gives good accuracy; hence we pre-train the classifier \CA{on 3.5M collected spectrogram-category pairs} \KG{at a variety of positions in the training environments} and freeze it during policy training. 
In contrast, location prediction %
\KGmon{is learned better on-policy.}
Training the $L_t$ \Z{predictor} on-policy \KGcr{has} %
the benefit of matching the training data distribution with policy behavior, leading to higher accuracy \KG{(see Supp.)}. \CA{We use the same experience collected for policy training to train the location predictor and update them at the same frequency.}
We use the mean squared error loss for the location predictor and the cross entropy loss for the goal \KG{object} label predictor.}

For policy training, we follow a two-stage training paradigm (as shown to be effective for transformer-based models \cite{silvio-scene-memory}) using decentralized distributed proximal policy optimization (DD-PPO)~\cite{wijmans_dd-ppo_2020}. 
In the first stage, we set the memory size $s_M=1$ (the most recent observation) \CCA{to train the observation encoder without attention.}
Then, in the second stage, we freeze the observation encoder and train the rest of the model \KG{with the full memory size} (\KGcr{$s_M=150$}).   \CA{In both stages, the loss consists of a value network loss \CA{to reduce the error of state-value prediction}, a policy network loss \CA{to produce better action distributions}, and an entropy loss to encourage exploration. We refer readers to PPO~\cite{schulman2017proximal} for more details. }  
\KG{To train the policy, we} reward the agent with $+10$ if it reaches the goal successfully and \KG{issue} an intermediate reward of $+1$ for reducing the geodesic distance to the goal, and an equivalent penalty for increasing it. We also issue a time penalty of $-0.01$ per time step to encourage efficiency. 

\ZA{To avoid sampling easy episodes (e.g., short or straight-line paths), we} require the geodesic distance \ZA{from the start pose} to the goal \ZA{to be} greater than 4 m and the ratio of Euclidean distance to geodesic distance to be greater than 1.1.} %
We collect 0.5M/500/1000 episodes for train/val/test splits \KG{for all 85 Matterport3D SoundSpaces environments.}

We train our model with Adam~\cite{kingma2014adam} with a learning rate of $\num{2.5e-4}$ for the policy network and $\num{1e-3}$ for the descriptor network.
\CA{We roll out policies for $150$ steps, %
update them \cc{with collected experiences} for two epochs, and repeat this procedure until convergence.
We train all methods, both ours and the baselines, for $300$M steps for them to fully converge.

At each time step, the agent receives a binaural audio clip of 1s, represented as \KG{two} $65 \times 26$ spectrograms.  \KGmon{The audio is computed by convolving the appropriate impulse response from SoundSpaces with the source audio waveform, thereby generating the sound the agent would hear in that environment at \KGcr{its current} position relative to the source.}
The RGB and depth images are center cropped to $64 \times 64$.
Both the observation encoder CNNs \KG{$f_B$ and $f_I$} and the descriptor network \KG{$f_D$} use a simplified ResNet-18~\cite{he_deep_2015} that is trained from scratch.
For the transformer model, we use one encoder layer and one decoder layer, \CA{which employ multi-attention with 8 heads.} 
The hidden state size is $256$ and the memory size $s_M$ is $150$, \CA{matching the frequency of policy updates}. 
}

\begin{table*}[t!]
\setlength{\tabcolsep}{2pt}
\centering
\begin{tabular}{ c| SSSSS | S S S S S}
 \toprule
                                                & \multicolumn{5}{c|}{\textit{Heard Sounds}} & \multicolumn{5}{c}{\textit{Unheard Sounds}} \\
                                                & {Success $\uparrow$} & {SPL $\uparrow$} & {SNA $\uparrow$} & {DTG $\downarrow$} & {SWS $\uparrow$}  & {Success $\uparrow$} & {SPL $\uparrow$} & {SNA $\uparrow$} & {DTG $\downarrow$} & {SWS $\uparrow$}\\
 \midrule
    Random                                      & 1.4  & 3.5  & 1.2 &  17.0  & 1.4    & 1.4  & 3.5 & 1.2  &  17.0  & 1.4\\
    ObjectGoal RL                               & 1.5  & 0.8  & 0.6 &  16.7  & 1.1    & 1.5  & 0.8  & 0.6 &  16.7  & 1.1\\
    \CAcr{Gan et al.~\cite{gan2019look}}         & 29.3  & 23.7 & \bfseries 23.0  & 11.3  & 14.4   & 15.9 & 12.3  & 11.6 &  12.7  & 8.0\\
    Chen et al.~\cite{chen_audio-visual_2019}   & 21.6  & 15.1 & 12.1  & 11.2 & 10.7    & 18.0  & 13.4 & 12.9  & 12.9  & 6.9 \\
    AV-WaN~\cite{chen_waypoints_2020}           & 20.9  & 16.8 & 16.2  & 10.3  & 8.3  & 17.2  & 13.2 & 12.7  & 11.0 & 6.9 \\
    SMT~\cite{silvio-scene-memory} + Audio      & 22.0  & 16.8 & 16.0  & 12.4  & 8.7  & 16.7  & 11.9 & 10.0  & 12.1  & 8.5 \\
    SAVi  (\textbf{Ours})                     & \bfseries 33.9  & \bfseries 24.0 &  18.3 & \bfseries 8.8  & \bfseries 21.5  & \bfseries 24.8  & \bfseries 17.2 &  \bfseries 13.2  & \bfseries 9.9  & \bfseries 14.7 \\
 \bottomrule
\end{tabular}
\caption{Navigation performance \ZA{on the SoundSpaces Matterport3D dataset~\cite{chen_audio-visual_2019}. Our SAVi model has higher success rates and follows a shorter trajectory (SPL) to the goal compared to the state-of-the-art. Equipped with its explicit goal descriptor \KG{and having learned semantically grounded object sounds from training environments,} our model is able to reach the goal more efficiently---even after it stops sounding---at a significantly higher rate than the closest competitor (\KGcr{see the SWS metric}).}}

\label{tab:main_results}
\end{table*}

\section{Experiments}

\paragraph{Baselines.}
We compare our model to the following baselines and existing work:
\begin{enumerate}[leftmargin=*,itemsep=0pt]
    \item \textbf{Random}: A random baseline that uniformly samples one of three actions and executes \textit{Stop} automatically when it reaches the goal (perfect stopping). 
    \item \textbf{ObjectGoal RL}: \CA{An end-to-end RL policy \ZA{with a GRU encoder and RGB-D inputs (no audio).} It is given the one-hot encoding of the \KG{true} \ZA{category} label as \ZA{an additional} input to search for the \ZA{goal} object instance. 
    This baseline is widely used in ObjectGoal tasks~\cite{gupta2017cognitive,chaplot_object_2020,mousavian_visual_2018,chang2020semantic}. 
        We train this method for 800M steps \CAcr{with perfect stopping}.  \KGmon{Details in Supp.}}
    \item \textbf{Gan et al.~\cite{gan2019look}}: \CAcr{A modular \KGcr{audio-visual} model that trains a goal location predictor offline and uses a geometric planner for planning. Since the original 
    \ZAcr{model} can not handle sporadic audio events,  we improve \KGcr{its} goal location predictor with our update operation $f_{\lambda}$.}
    \item \textbf{Chen et al.~\cite{chen_audio-visual_2019}}: An end-to-end RL policy that encodes past memory with a GRU \KG{RNN} and is trained to reach the goal using audio and visual observations.
    \item \textbf{AV-WaN~\cite{chen_waypoints_2020}}: A hierarchical RL model that \KG{records acoustic observations on the ground plane,} predicts waypoints, and uses a path planner to move towards these waypoints using a sequence of navigation actions. 
    \item \textbf{SMT~\cite{silvio-scene-memory} + Audio}: \CA{We adapt the \KG{scene memory transformer} (SMT) model~\cite{silvio-scene-memory} to our task by also encoding the audio observation in \Z{its}  
    memory.}
    \KG{Unlike our model, it} does not explicitly predict the goal description and relies only on the cues available in memory to reach the goal. %
    The \CAcr{latest} observation embedding is used \CCA{as decoder input} to decode $M_e$ and predict the state. %
\end{enumerate}

\CA{All models use the same reward function \CCA{and inputs}.}
\KG{For all methods, there is no actuation noise since audio rendering is only available at grid points \Z{(see~\cite{chen_audio-visual_2019} for details).}}

\vspace{-0.15in}
\paragraph{Metrics.} We evaluate the following navigation metrics: 1) success rate: the fraction of successful episodes; 2) success weighted by inverse path length (SPL): the standard metric~\cite{anderson2018evaluation} that weighs successes by their adherence to the shortest path; 3) success weighted by inverse number of actions (SNA)~\cite{chen_waypoints_2020}: this penalizes collisions and in-place rotations by counting number of actions instead of path lengths; 4) average distance to goal (DTG): the agent's distance to the goal when episodes are finished; 5) success when silent (SWS): the fraction of successful episodes when the agent reaches the goal after the end of the acoustic event.

\begin{figure*}[hbt!]
    \centering
    \includegraphics[width=1.\textwidth]{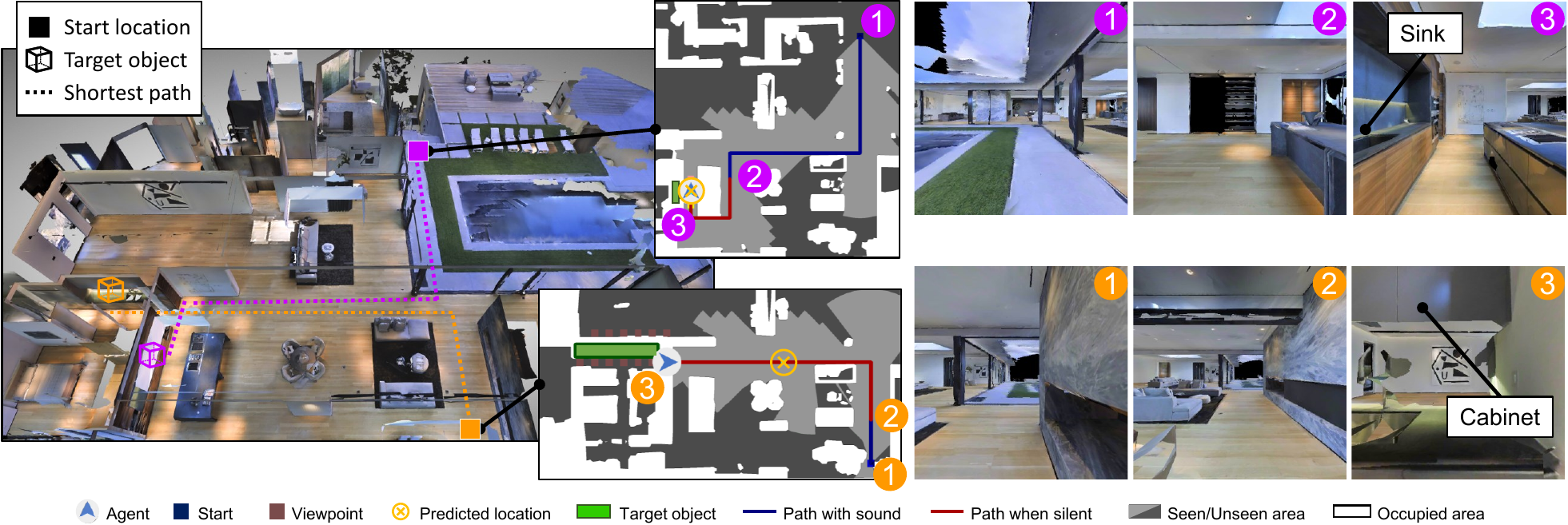}
    \caption{
    \KGcr{Example SAVi navigation trajectories.}  %
    \Z{In the first episode (top/magenta) the agent hears a water dripping sound and in the second episode (bottom/orange) a sound of opening and closing a door.
    For each episode, we show three egocentric visual views (right) sampled from the agent's trajectory at the start location~\raisebox{.5pt}{\textcircled{\raisebox{-.9pt} {1}}}, when the sound stops~\raisebox{.5pt}{\textcircled{\raisebox{-.9pt} {2}}}, and at the end location~\raisebox{.5pt}{\textcircled{\raisebox{-.9pt} {3}}}.
   In the top episode, the acoustic event lasts for two thirds of the trajectory and when the sound stops the agent has an accurate estimate of the object location that helps it find the sounding object (the sink).
    The second episode (bottom) has a much shorter acoustic event. The agent's estimate of the object location is inaccurate when the sound stops but still helps the agent as a general directional cue. 
    The agent leverages this spatial cue \emph{and} the semantic cue from its estimate of the object category, a cabinet, to attend to its multimodal memory to find the object in the kitchen and end the episode successfully.} 
    }
    \label{fig:trajectories}
\end{figure*}

\vspace{-0.15in}
\paragraph{Navigation results.} 
\ZA{\KG{Following} \KGcr{standard} protocol~\cite{chen_audio-visual_2019}} we \ZA{evaluate all models in} two settings: 1) \textit{heard sounds}---train and test on the same sound  2) \textit{unheard sounds}---train and test on disjoint sounds. In both cases, the test environments are always unseen, hence both require generalization. 
\CA{All results are averaged over 1,000 test episodes.}

Table~\ref{tab:main_results} shows the results. %
Our \KGcr{SAVi} approach outperforms all other models by a large margin on all metrics---with \CAcr{0.3\%,} 8.9\%, 7.2\%, 7.2\% \KG{absolute gains in} SPL on \textit{heard sounds} and \CAcr{4.9\%,} 3.8\%, 4\%, 5.3\% \ZAcr{absolute} SPL gains on \textit{unheard sounds} compared to \CAcr{Gan et al.~\cite{gan2019look},} Chen et al.~\cite{chen_audio-visual_2019}, AV-WaN~\cite{chen_waypoints_2020}, and SMT~\cite{silvio-scene-memory}, respectively. 
This shows our model leverages audio-visual cues intelligently and navigates to goals more efficiently.  %
AV-WaN represents the state-of-the-art for AudioGoal audio-visual navigation.  Our SAVi model's gains over AV-WaN show both 1) the distinct new challenges offered by the semantic AudioGoal task, and 2) our model's design effectively handles them.\footnote{\KG{
\ZA{While }AV-WaN~\cite{chen_waypoints_2020} reports large performance improvements over Chen et al.~\cite{chen_audio-visual_2019} on \ZA{the standard} AudioGoal \ZA{task}, we do not \ZA{observe similar margins between the two models} here. We attribute this to temporal gaps in the memory caused by AV-WaN's waypoint formulation---which are not damaging for constantly sounding targets, but do cause problems for semantic AudioGoal (see Supp.~for details).}}

In addition, our model improves the success-when-silent (SWS) metric by a large margin \ZA{compared to the closest competitor.}
\ZA{This} emphasizes the advantage of our goal descriptor \ZA{module}.
\ZA{The explicit and persistent descriptor for the goal in our model helps to maintain the agent's focus on the target even after it stops emitting a sound.
Although the SMT+Audio~\cite{silvio-scene-memory} \KG{model also has} access to a large memory pool and can leverage implicit goal information from old observations, \KG{lacking our goal descriptor and the accompanying \ZA{goal-driven} attention, 
} it underperforms our model by a sizeable margin.
}

As expected, Random does poorly on this task due to the challenging complex environments. 
\CA{\ZA{Although} ObjectGoal RL has the goal's ground truth category label \ZA{as input, it fails in most cases.} %
\ZA{This shows that knowing the category label by itself is insufficient to succeed in this task; the agent needs to locate the specific instance of that category, which is difficult without the acoustic cues.}}

\vspace{-0.15in}
\paragraph{Navigation trajectories.} 
Figure~\ref{fig:trajectories} shows \ZA{test episodes for our SAVi model}.  
The agent uses its \Z{acoustic-visual} perception \Z{and memory} along with the spatial and semantic cues from the acoustic event\Z{, whether from a long event (water dripping sound) or a short one (opening and closing a door sound),} to successfully find the target objects (the sink and the cabinet).  %
\KG{See the Supp.~video for more examples.}  

\KGcr{Common failure cases are when: 1) the sound stops too early in the episode, and the agent has not accumulated enough spatial or semantic cues about the goal. In this case the agent might either search for the wrong object (noisy semantics) or search for the object in the wrong place (noisy location); 2) the agent issues a premature stop action near the target object but not exactly at the right location.}

\vspace{-0.15in}
\paragraph{Distractor sounds.} 

\KG{In our tests so far, there is a single acoustic event per episode, whether comprised of a heard or unheard sound (Table~\ref{tab:main_results}). Next, we generalize the setting further to include \emph{unheard distractor sounds}---sounds happening simultaneously with the target object. This corresponds to real-world scenarios, for example, where the door slams shut while the AC is humming. 
For this setting to be well-defined, the agent must know which sound is its target; hence, we input the one-hot encoding of the target object to all models \CA{and concatenate it with their state features. 
For our model, in addition to replacing $C_t$ with this one-hot encoding, \ZA{we also use it as input to the location prediction network along with $B_t$.}
This \ZA{allows the} location prediction network to \KG{learn to} \ZA{identify which of the sounds mixed in the input needs to be localized.} %
}
\CA{We use the 102 periodic sounds from SoundSpaces~\cite{chen_audio-visual_2019} as the \Z{set of possible} distractor sounds, \KG{which are disjoint from the target object sounds curated for this work.
\C{We divide these 102 sounds into non-overlapping 73/11/18 splits for train/val/test, and hence the distractor sound at test time is unheard.}}
In each episode, we randomly position one distractor sound in the environment \KG{at a location different from the goal}.
}}

Table~\ref{tab:distractor_sounds} shows the results. %
While the performance of the baselines \KG{suffers} from the distracting environment noise, our agent is still able to reach a success rate of $11.8\%$ and SPL of $7.4\%$, which is $7.6\%$ and $4.5\%$ higher than the best-performing baseline. %
\CA{This shows the proposed inferred goal descriptor helps the agent attend to important observations to capture semantic and spatial cues, making our model more robust to the environment noise.}
\KGmon{That said, the absolute performance declines for all methods in this hard setting.  We plan to investigate ways to explicitly separate the ``clutter" sounds in future work.}

\begin{table}[t!]
\setlength{\tabcolsep}{2pt}
\centering
\begin{tabular}{ c|S S S S S}
 \toprule
                                                & {Success $\uparrow$} & {SPL $\uparrow$} & {SNA $\uparrow$} & {DTG $\downarrow$} & {SWS $\uparrow$}\\
 \midrule
    Chen et al.~\cite{chen_audio-visual_2019}   & 4.0   & 2.4  & 2.0 & 14.7   & 2.3  \\
    AV-WaN~\cite{chen_waypoints_2020}           & 3.0   & 2.0  & 1.8 & 14.0   & 1.6   \\
    SMT~\cite{silvio-scene-memory}+Audio        & 4.2   & 2.9  & 2.1 & 14.9   & 2.8   \\
    SAVi  (\textbf{Ours})                       & \bfseries 11.8  & \bfseries 7.4 & \bfseries 5.0  & \bfseries 13.1  & \bfseries 8.4 \\
 \bottomrule
\end{tabular}
\caption{Navigation performance on \textit{unheard sounds} \ZA{in the presence of \CA{unheard} distractor sounds.}}
\label{tab:distractor_sounds}
\end{table}

\vspace{-0.15in}
\paragraph{\ZA{Analyzing the goal descriptor}.} 
Next we ablate the \ZA{two main components} in the goal descriptor, location and category, to study their relative impact \CA{for the \textit{unheard sounds} experiment from Table~\ref{tab:main_results}}. %
Table~\ref{tab:ablations} shows that ablating any component results in a performance drop. $L_t$ \ZA{has a} comparatively %
\ZA{larger impact on our model's performance.}

\KG{Next} we analyze the \emph{successful} episodes in the context of $L_t$ and $C_t$.
For 56\% of them, our model \ZA{ends the episode by stopping at its own estimate of the goal location in its descriptor,} \KG{suggesting that the agent has successfully used its directional sound prediction to guide its movements.} %
\ZA{On the other hand, for the other 44\%, the agent stops at a \KG{(correct)} location \emph{different} than $L_t$,} \KG{suggesting that the agent has relied more on the visual context cues leading to the anticipated object $C_t$.}
\KG{In fact, if we inject a random category label instead of $C_t$ at the start of the episode, success rates and SPL drop up to $8\%$.}
\ZA{The learned associations between the spatial and semantic cues are %
important for success; breaking these associations with random category labels forces the agent to attend to contradictory cues about the goal in its memory, thus increasing the chance of failure.}

\CA{To understand if the performance gain comes from our goal descriptor or the transformer, we \KGcr{further ablate} our model by replacing the transformer with \KGcr{an} RNN. \ZAcr{We find that our goal descriptor network also provides significant improvements when combined with RNNs} (see Supp.).}

\begin{table}[t!]
\setlength{\tabcolsep}{2pt}
\centering
\begin{tabular}{ c| S S S S S}
 \toprule
                                & {Success $\uparrow$} & {SPL $\uparrow$} & {SNA $\uparrow$} & {DTG $\downarrow$} & {SWS $\uparrow$}\\
 \midrule
    $C_t$-only                   & 20.5  & 13.5  & 11.6  & 9.8  & 11.0 \\
    $L_t$-only                   & 23.9  & 16.2  & \bfseries 13.5  & \bfseries 9.3  & 13.8 \\
    w/o  aggregation            & 21.9  & 14.3  & 11.1  & 9.7  & 13.4 \\
    Full model                  & \bfseries 24.8  & \bfseries 17.2 &  13.2  &  9.9  & \bfseries 14.7 \\
 \bottomrule
\end{tabular}
\caption{Ablation experiment results.
} 
\label{tab:ablations}
\end{table}

\vspace{-0.15in}
\paragraph{Goal descriptor \KG{accuracy and aggregation}.} 
The goal descriptor network has two main modules: 1) $f_D(\cdot)$, which produces the current descriptor estimate and 2) an aggregation function $f_\lambda(\cdot)$, which aggregates the current estimate with the previous goal descriptor. Next we evaluate goal prediction accuracy with and without aggregation, as well as how aggregation impacts the navigation performance.

\KG{The average location prediction error is $8.1$ m and the average category prediction accuracy is $64.5\%$ with aggregation, and $8.4$ m, $53.6\%$ without aggregation.} 
Aggregation is \CA{important because the source sound is divided into 1s clips for each step, and the characteristics of the sound in some seconds are harder to identify, e.g., the silent moment between pulling and pushing a chest of drawers.}
\KG{Essentially, aggregation} stabilizes the goal descriptor prediction. 
\cc{See Supp.~for the distribution of prediction accuracy over distance to goal.}
\KG{Navigation performance is affected as well: success rate and SPL drop about 3 points without aggregation
(``w/o aggregation" ablation in Table~\ref{tab:ablations}).}

\begin{figure}
    \centering
    \includegraphics[width=0.48\textwidth]{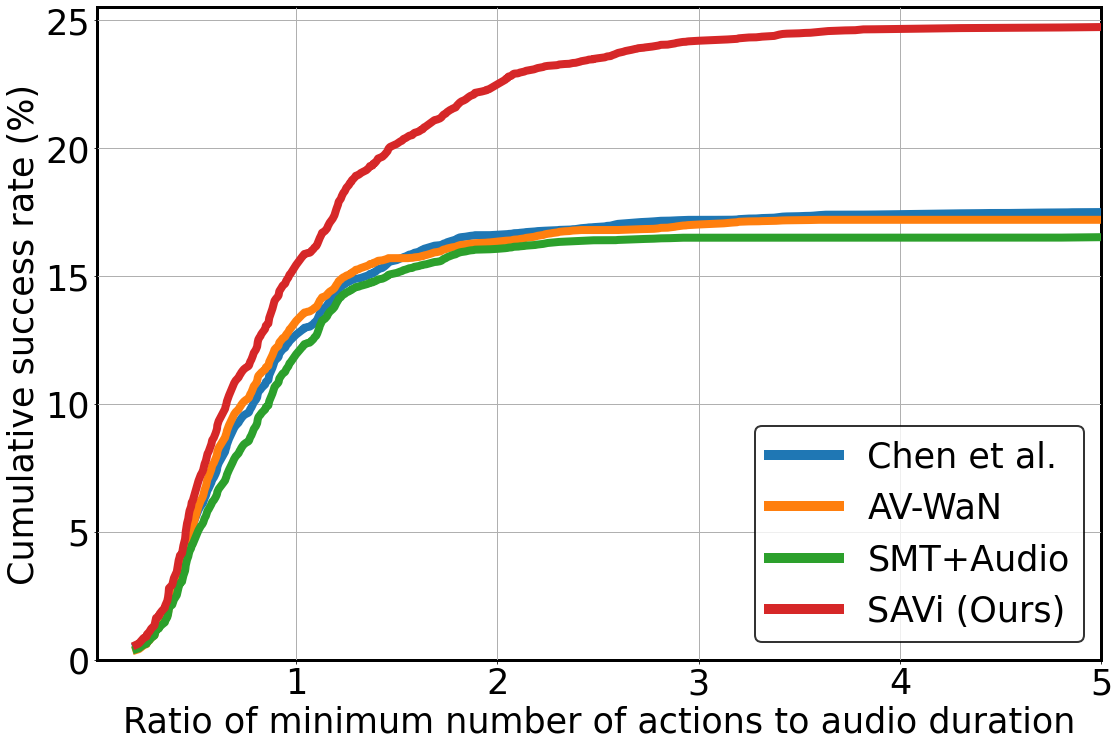}
    \caption{Cumulative success rate vs.~silence percentage.
    }
    \label{fig:success_vs_silence}
\end{figure}

\vspace{-0.15in}
\paragraph{\ZA{Robustness to silence duration}.}  %
Figure~\ref{fig:success_vs_silence} analyzes how the models perform after the \KG{goal} sound stops.  We plot the cumulative success rate vs.~silence \KG{ratio}, where the latter is the ratio of the minimum number of actions \KG{required to reach the goal} to the duration of audio. A point $(x,y)$ on this plot means the \CCA{fraction of successful episodes with ratios up to $x$ among all episodes is $y$.}
When this ratio is greater than 1, no agent can reach the goal before the audio stops.  
The greater this ratio is, the \KG{longer the fraction} of silence, \KG{and hence the harder the episode.} %
Indeed, we see for all models the success rate accumulates more slowly as the ratio becomes bigger.
However, while the success rates of Chen et al.~\cite{chen_audio-visual_2019}, AV-WaN~\cite{chen_waypoints_2020}, and SMT~\cite{silvio-scene-memory} increase only marginally for ratios greater than 1, our model shows a noticeable increase after the ratios surpass 1 and even 2.
\KG{This indicates} our model is able to cope with long silence to reach goals, \KG{thanks to the guidance of} our predicted goal descriptor \KG{and its attention on the memory}.

\section{Conclusions}
\CA{We introduce the task of semantic audio-visual navigation in complex 3D environments. To support this task, we expand an existing audio simulation platform to provide semantically grounded object sounds. We introduce a transformer-based model that learns to predict a goal descriptor capturing both spatial and semantic properties of the target. By encoding the observations conditioned on this goal descriptor, our model learns to associate acoustic events with visual observations. 
We show that our approach outperforms existing state-of-the-art models. 
\C{We provide an in-depth analysis of the impact of the goal descriptor and its components, and show that our model is more robust to long silence duration and acoustic distractors.}
\KG{In future work,} we are interested in generalizing policies learned in \KG{these high quality simulators} to test in the real world.}

\vspace*{0.2in}
\noindent\textbf{Acknowledgements:}
UT Austin is supported in part by \KGcr{the IFML NSF AI Institute and DARPA Lifelong Learning.  Thanks to Santhosh Ramakrishnan for helpful discussions.}

\clearpage

{\small
\bibliographystyle{ieee_fullname}
\bibliography{main}
}

\clearpage

\section{Supplementary Material}
In this supplementary material, we provide additional details about:
\begin{enumerate}[leftmargin=*,itemsep=0pt]
    \item Video (with audio) for qualitative assessment of our agent’s performance. \KG{Please listen with headphones to hear the spatial sound properly.}
     
    \item Implementation details and analysis \ZAcr{of the baselines} (Sec. 5)
    
    \item \CAcr{Ablation of \ZAcr{the policy network}}
    
    \item Distribution of prediction accuracy over distance to goal (Sec. 5)
    
    \item \CAcr{Analysis of semantic audio-visual navigation with distractors}
    
    \item On-policy location predictor training (Sec. 4.4)
    
    \item \CAcr{Ablation with true goal category and location}
\end{enumerate}

\subsection{Qualitative Video}
The supplementary video\footnote{\url{https://youtu.be/EKCYc1dFOhw}} demonstrates the audio simulation platform that we use and shows the comparison between our proposed model and the baselines as well as qualitative analysis for failure cases. 
Please listen with headphones to hear the binaural audio correctly.

\subsection{Implementation Details and Analysis \ZAcr{of the Baselines}}
\paragraph{ObjectGoal RL.} 
We implement this baseline by first feeding the RGB-D \ZA{observations} into a CNN \ZA{(similar CNN to $f_I(\cdot)$ in our model)} and concatenating the visual features with a one-hot encoding of the target label. 
A one-layer GRU memory takes the concatenated feature as input and outputs a state vector of size 512. 
Similar to our work, this state representation is used by an actor-critic network to predict the action distribution and value of the current state. 
\ZAcr{Furthermore, we use perfect stopping for this baselines since the model performs poorly with a learned stop action.}

Although this baseline has the goal's ground truth label \CAcr{and perfect stopping}, it fails \ZA{quite often in reaching the goal}. 
This shows knowing the category alone is insufficient to locate the particular object instance and to succeed in this task.  The model needs to leverage \ZA{both visual and acoustic} cues \KG{to find} the \ZA{goal}. %
\KG{This experiment also draws attention to the difference between the proposed semantic AudioGoal and the existing task of ObjectGoal.}

\paragraph{Gan et al.~\cite{gan2019look}} %
\KG{We compare to} \ZA{the model from} Gan et al.~\cite{gan2019look}
, which trains a \ZA{goal} location predictor in an offline fashion and uses a geometric planner for planning a path to the predicted \ZA{goal} location. %
We use the same amount of training data for our category predictor to train the goal predictor \ZA{from \cite{gan2019look}}. 
\ZA{The original model from~\cite{gan2019look} assumes a continuous periodic acoustic event and it cannot handle sporadic or short acoustic events like those considered in this work. 
To \KG{improve the existing} model to perform in this task, we augment its goal location predictor with our}
update operation $f_{\lambda}$ (Sec. 4.2) %
for transforming the predicted location \ZA{when the audio goal becomes silent}. 

\ZA{In evaluation, our observations confirm those} reported in \cite{chen_waypoints_2020}.
\ZA{Since the model does not leverage visual cues for reasoning \KG{about the goal location}, it does not learn to associate visual and acoustic cues with scene observations and goal properties. Therefore, it is more }prone to errors and \ZA{the agent} suffers  from backtracking \ZA{its steps quite often} when the \ZAcr{goal location} prediction is inaccurate. 
\ZA{The model} achieves 15.9\% success rate and 12.3\% SPL on the \textit{unheard sound} test split\ZA{, compared to our SAVi model 24.8\% success rate and 17.2\% SPL}. 
\ZA{While \cite{gan2019look}} leverages external supervision for training the location predictor, this is not enough to solve this task \ZA{efficiently} because the agent needs to fully leverage the semantic and spatial cues from audio along with its visual perception to locate the sounding objects.

\paragraph{AV-WaN~\cite{chen_waypoints_2020}.}
While AV-WaN~\cite{chen_waypoints_2020} reports large performance improvements over Chen et al.~\cite{chen_audio-visual_2019} on the standard AudioGoal task \ZA{(see~\cite{chen_waypoints_2020} for details)}, we do not observe similar margins between the two models here. 
Both models\ZA{, AV-WaN~\cite{chen_waypoints_2020} and Chen et al.~\cite{chen_audio-visual_2019},} use RNNs to encode the state representation; however, AV-WaN accumulates the observations at the waypoint prediction level while Chen et al. does so at each step.
We speculate that this behavior creates large temporal gaps in the memory for AV-WaN, which makes it harder for the model to adapt to the more challenging task of semantic AudioGoal; the sound may stop at any moment, and the AV-WaN model may not be able to capture the last important acoustic cues in between waypoints. 
Our model outperforms both since it can keep track of a large set of observations and leverage this information at each step while navigating.

\subsection{Ablation of \ZAcr{the Policy Network}}
\ZAcr{To analyze the impact of the transformer architecture of the policy network on our model performance,}
\CAcr{
we include an additional ablation by replacing the transformer \ZAcr{in our SAVi model} with a typical RNN+MLP for action and value prediction (similar to \ZAcr{\cite{chen_audio-visual_2019}}).
Table~\ref{tab:ablations} shows the results in the unheard sounds setting.
We see that a significant part of the performance improvement comes from our goal descriptor network (GDN) contribution. 
Both models \ZAcr{(SAVi w/ Transformer and SAVi w/ RNN+MLP)} benefit significantly from having our GDN, with the transformer model leading to best performance since it allows the GDN to attend to longer observation sequences compared to the RNN. 

\begin{table}[t]
\setlength{\tabcolsep}{2pt}
\centering
\scalebox{0.8}{
\begin{tabular}{ l SSSSS}
 \toprule
                                & {Success $\uparrow$} & {SPL $\uparrow$} & {SNA $\uparrow$} & {DTG $\downarrow$} & {SWS $\uparrow$}\\
 \midrule
    \multicolumn{6}{l}{\ZAcr{\textbf{RNN Policy Network}}}\\
    Chen et al.~[11]                 & 18.0  & 13.4 & \bfseries 12.9  & 12.9  & 6.9 \\
    SAVi w/ RNN+MLP (Ours)           & \bfseries 21.4  & \bfseries 15.4  &  12.6  &  \bfseries 9.8  & \bfseries 11.2 \\
\midrule
    \multicolumn{6}{l}{\ZAcr{\textbf{Transformer Policy Network}}}\\
    SMT~[15] + Audio                 & 16.7  & 11.9 & 10.0  & 12.1  & 8.5 \\
    SAVi w/ Transformer (Ours)       & \bfseries 24.8   &  \bfseries 17.2 &  \bfseries 13.2  &  \bfseries 9.9  &  \bfseries 14.7 \\
 \bottomrule
\end{tabular}
}
\caption{Ablation \ZAcr{of our policy network} with a typical RNN+MLP.}
\label{tab:ablations}
\end{table}
}

\subsection{Distribution of \ZA{Goal Descriptor} Accuracy}
\begin{figure}
    \centering
    \subcaptionbox{}{\includegraphics[width=0.225\textwidth]{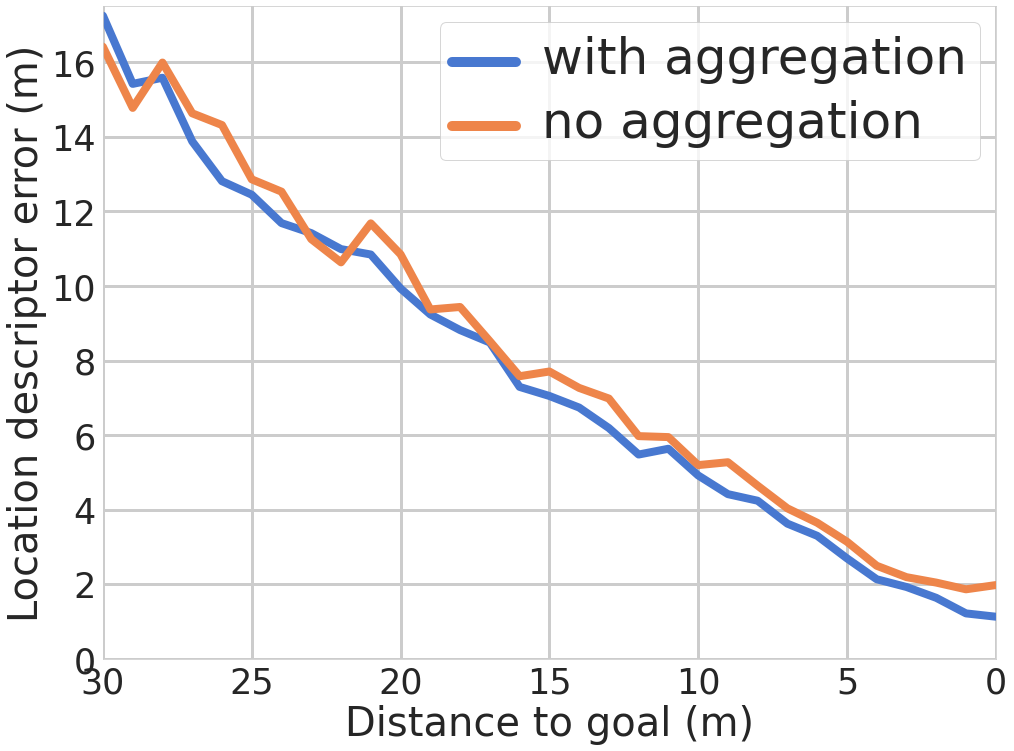}}\hfill
    \subcaptionbox{}{\includegraphics[width=0.225\textwidth]{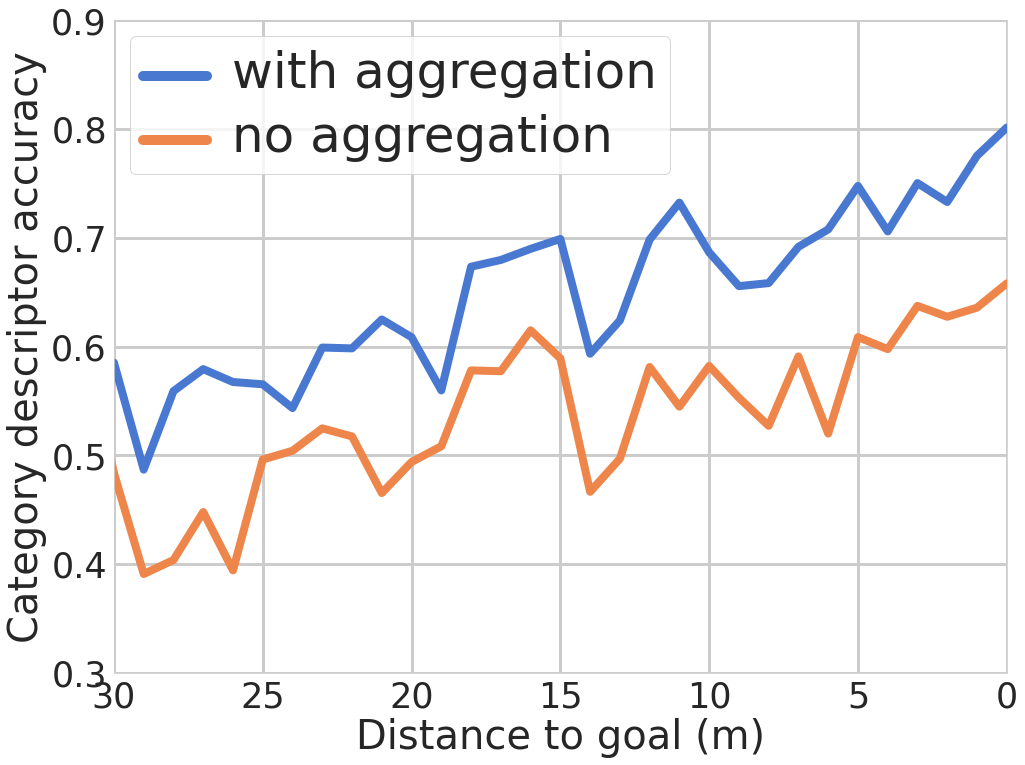}}
    \caption{Error analysis of the location predictions and the category predictions \ZA{in the goal descriptor} as a function of \CA{the agent's geodesic distance to goal}.
    }
    \label{fig:prediction_accuracy}
\end{figure}

Figure~\ref{fig:prediction_accuracy} shows how the location descriptor error and the category descriptor \ZA{accuracy} change as the agent gets closer to the goal with and without temporal aggregation. 
The location error is measured as the Euclidean distance between the predicted %
and the ground truth goal location. 
The category accuracy is measured by \KG{whether} the correct goal is predicted or not. 
We can see that the \KG{error} of both predictions get \KG{lower} as the agent gets closer to the goal location \ZA{and the temporal aggregation leads to higher performance.} 

\subsection{Analysis of Semantic Audio-Visual Navigation with Distractors}
\ZAcr{We have evaluated in the main paper the navigation performance of our model in the presence of acoustic distractors.}
The target and distractor sounds are disjoint \ZAcr{in this setting} and both are unheard at test time, which poses a great challenge for the agent to clearly separate the mixed audio signal. 
We believe this is a main factor in the performance drop seen by all models, though ours remains best (Table~2 in the main paper).

\begin{table}[t]
\setlength{\tabcolsep}{5pt}
\centering
\scalebox{0.95}{
\begin{tabular}{l|ccccc}
 \toprule
            & Beeps     & Music     & Creak     & Horn      & Telephone\\
 \midrule
Chair	    &	0.26	&	0.28	&	0.20	&	0.20	&	0.24\\
Cabinet	    &	0.25	&	0.25	&	0.14	&	0.12	&	0.23\\
Counter	    &	0.28	&	0.47	&	0.34	&	0.25	&	0.41\\
Sink	    &	0.03	&	0.07	&	0.03	&	0	    &	0.07\\
TV      	&	0.14	&	0.19	&	0.19	&	0.14	&	0.19\\
 \bottomrule
\end{tabular}
}
\caption{Success rate of goals (rows) in the presence of various distractors (columns).
\ZA{We test our model with a single distractor type in each test run, and normalize the SR by the number of episodes for each goal type.}
}
\label{tab:distractors}
\end{table}

\ZAcr{To further analyze the impact of acoustic dsitractors,} we conduct an ablation of our model by changing the type of distractors \ZAcr{at each test run}.
Table~\ref{tab:distractors} shows a subset of the (goal, distractor) combinations.
Indeed, when the distractor sound is sufficiently different from the goal (e.g., Music and Telephone), the model performs well, but  when it is similar (e.g., Cabinet and Creak) or much louder (e.g., Horn) then it is harder for the model to extract a clear signal for the goal.

\subsection{On-policy Location Predictor Training}
\KG{As noted in the main paper,} we find training the location predictor on-policy and online leads to higher accuracy compared to using a pretrained model. %
If we use \ZA{an off-policy model in our approach (i.e., similar to }the location predictor trained for Gan et al.), this version underperforms our model by 4.8\% success rate and 4.7\% SPL on the \textit{unheard sound} test split.

\vspace{0.1in}
\subsection{Ablation with True Goal Category and Location}
\ZAcr{Our SAVi model learns to predict the goal descriptor (i.e., location and category) based on the heard acoustic cues while navigating.}
\ZAcr{To show an upper bound performance for our goal descriptor network, }\CAcr{we supply the model with the true goal category and location instead of \ZAcr{the predictions}. Our model achieves} 65\% SPL \ZAcr{compared to the 24\% SPL under the same setting but with predicted descriptors}.
Note that when the ground truth location of the goal is available at each step, 
the task boils down to the common PointGoal navigation~\cite{habitat19iccv,Chaplot2020Learning}. 

\end{document}